\documentclass[review]{elsarticle}
\usepackage{lineno,hyperref}
\usepackage{amsmath}
\usepackage{booktabs}
\usepackage{threeparttable}
\usepackage{multirow}
\usepackage{adjustbox}
\usepackage{rotating}
\usepackage{amsfonts}
\usepackage{graphicx}
\usepackage{wrapfig}
\usepackage{bm}
\usepackage[ruled,vlined]{algorithm2e}
\usepackage{subfig}
\modulolinenumbers[5]

\journal{Journal of \LaTeX\ Templates}

\begin{document}

\begin{frontmatter}

\title{Enhancing Visual Representation for Text-based Person Searching}

\author[MVTI]{Wei Shen}
\author[MVTI]{Ming Fang}
\author[MVTI]{Yuxia Wang}
\author[MVTI]{Jiafeng Xiao}
\author[MVTI]{Diping Li}
\author[MVTI]{Huangqun Chen}
\author[zjit,zju]{Ling Xu}
\author[jxuniversity]{Weifeng Zhang\corref{mycorrespondingauthor}}
\cortext[mycorrespondingauthor]{Corresponding author: Weifeng Zhang, zhangweifeng@zjxu.edu.cn}


\address[MVTI]{Metrological Verification and Testing Institute of Jiaxing, China}
\address[zjit]{School of Artificial Intelligence, Zhejiang Sci-Tech University, Zhejiang, China}
\address[zju]{College of Computer Science and Technology, Zhejiang University, Zhejiang, China}
\address[jxuniversity]{College of Information Science and Engineering, Jiaxing University, Zhejiang, China}

\begin{abstract}
Text-based person search aims to retrieve the matched pedestrians from a large-scale image database according to the text description. The core difficulty of this task is how to extract effective details from pedestrian images and texts, and achieve cross-modal alignment in a common latent space. Prior works adopt image and text encoders pre-trained on unimodal data to extract global and local features from image and text respectively, and then global-local alignment is achieved explicitly. However, these approaches still lack the ability of understanding visual details, and the retrieval accuracy is still limited by identity confusion. In order to alleviate the above problems, we rethink the importance of visual features for text-based person search, and propose \textbf{VFE-TPS}, a \textbf{V}isual \textbf{F}eature \textbf{E}nhanced \textbf{T}ext-based \textbf{P}erson \textbf{S}earch model. It introduces a pre-trained multimodal backbone CLIP to learn basic multimodal features and constructs Text Guided Masked Image Modeling task to enhance the model's ability of learning local visual details without explicit annotation. In addition, we design Identity Supervised Global Visual Feature Calibration task to guide the model learn identity-aware global visual features. The key finding of our study is that, with the help of our proposed auxiliary tasks, the knowledge embedded in the pre-trained CLIP model can be successfully adapted to text-based person search task, and the model's visual understanding ability is significantly enhanced. Experimental results on three benchmarks demonstrate that our proposed model exceeds the existing approaches, and the Rank-1 accuracy is significantly improved with a notable margin of about $1\%\sim9\%$. Our code can be found at \url{https://github.com/zhangweifeng1218/VFE_TPS}.
\end{abstract}


\end{frontmatter}


\section{Introduction}
\label{lab:introduction}

\begin{figure*}
\centering
	\subfloat[]{\includegraphics[width = 0.8\linewidth]{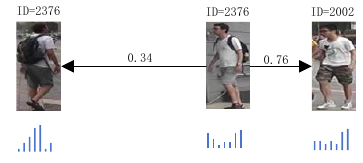}}
	\hspace{20mm}
	\subfloat[]{\includegraphics[width = 0.8\linewidth]{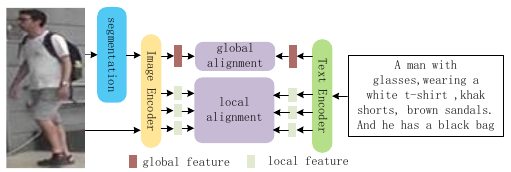}}
    \hspace{20mm}
	\subfloat[]{\includegraphics[width = 0.8\linewidth]{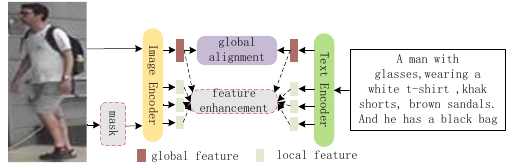}}
\caption{(a) Person identity confusion: The pedestrians with same ID have lower similarity than pedestrians with different IDs. The bar chart below each image represents its visual feature. (b) Existing global-local alignment approaches, which needs explicitly conduct global and local alignment for image and text. (c) Our visual feature enhancement aided matching paradigm, which introduces auxiliary tasks to enhance visual features and only global alignment is needed. The dashed arrows and boxes are only active during training.}
\label{fig:motivation}
\end{figure*}

Text-based person search \cite{R1} aims to retrieve the corresponding pedestrian images according to the text query which describes the attributes of the target person. It is a comprehensive application combining cross-modal retrieval \cite{R2}, person re-identification \cite{R3} and visual tracking \cite{vc1,vc2,vc3,vc4}. Different from traditional person re-identification task which uses images as query, text-based person search retrieves target pedestrian images based on text description, which is much more simple and intuitive. For example, when conducting a suspect search task, witnesses can only provide a text description of the suspect rather then a specific images. Therefore, text-based person search has better practical value, especially in the field of intelligent security. In the recent decay, text-based person search has become one of the hottest research topics in multimodal intelligence \cite{R4}, and has gained widespread attention from academia and industry.

While significant progress has been made in recent years, text-based person search task remains a challenging task. This is primarily due to: (1) The task of text-based person search is a typical cross-modal task, where the model needs to simultaneously understand the text query and the content of pedestrian images, extract appropriate features from both text and images, and then project them into a common latent space to achieve cross-modal alignment \cite{R5}. (2) In contrast to typical cross-modal retrieval tasks involving natural scene images, text-based person search encounters the challenge of subtle differences among person images, which leads to a common issue of person identity confusion \cite{R6}. As shown in Figure \ref{fig:motivation}(a), the left and middle images share the same person identity, yet their similarity is only 0.34, which is much lower than the similarity between the middle image and the right image which have different identities. This is mainly due to the fact that pedestrian images are often captured by different surveillance cameras at different times and locations, leading to variations in lighting, camera angles, and other environmental factors. The key to addressing the above problem lies in the algorithm's ability to extract detailed information, which is closely relevant to the current task, from the images. For example, in Figure \ref{fig:motivation}(a), the model needs to focus on details such as the color of the pedestrian's shoes and the presence of a backpack, rather than paying attention to factors like the pose of the pedestrian. Early solutions \cite{R7,R8,R9} rely on global feature alignment, which typically involve using convolutional neural networks and recurrent neural networks to extract global features from images and texts respectively, and then projecting them into a common latent space for alignment. These methods overlook local detailed features and have limited cross-modal interaction capabilities, resulting in low retrieval accuracy. Recognizing the significance of local features in text-based person search task, latter works \cite{R10,R11,R12,R13} adopt global-local alignment method, which further explore the extraction and alignment of local features on top of global alignment. As shown in Figure \ref{fig:motivation}(b), such methods typically utilize statistical priors for pedestrian images \cite{R14} or employ pre-trained segmentation models \cite{R15} to explicitly extract features of different image parts. Simultaneously, a natural language grammar parser \cite{R16} is used to parse the text query into phrases. Subsequently, alignment of image parts and text phrases can be conducted. While these global-local alignment approaches have made significant progress in retrieval accuracy compared to global alignment models, this explicit alignment of local features from images and texts is constrained by the capability of segmentation models and grammar parsers. This limitation results in inaccurate extraction of local features and aligning all local features individually during the inference stage leads to excessive computational costs. Furthermore, both the aforementioned global alignment and global-local alignment methods utilize image encoder and text encoder pre-trained on unimodal data, lacking the capability of multimodal feature learning.

In this paper, we propose a Visual Feature Enhanced Text-based Person Search (VFE-TPS) model, which is illustrated in Figure \ref{fig:motivation}(c). In order to extract rich cross-modal features from pedestrian images and text queries, we no longer rely on encoders pre-trained on unimodal data. Instead, we initialize our image and text encoders using Contrastive Language-Image Pre-training (CLIP) \cite{R17,n2} model which is pre-trained on a large-scale image-text dataset. Existing works have demonstrated CLIP's outstanding generalization and zero-shot capability in tasks such as image classification and video understanding \cite{n3}. However, the training images used by CLIP are natural scene images collected from the internet, which exhibit significant differences in feature distribution when compared to pedestrian images. Therefore, prior studies \cite{R18,R19} have indicated that directly transferring knowledge from CLIP to text-based person search task by freezing the model parameters is unfeasible. Therefore, instead of freezing the CLIP model parameters, we now use CLIP to initialize the image and text encoders. We design auxiliary training tasks to attempt the transfer and adaptation of the cross-modal knowledge embedded in CLIP to text-based person search.

Enhancing the model's ability of learning discriminative details is an essential approach to address text-based person search task. In this paper, we introduce two auxiliary tasks designed to enhance both global and local features without relying on prior statistical knowledge or extra annotations. Firstly, inspired by masked image modeling techniques \cite{R20,R21}, we design a Text Guided Masked Image Modeling (TG-MIM) task. In this task, we use text query as guidance to enhance task-related local features through a cross-attention mechanism. Subsequently, we fuse the local features using a self-attention mechanism and reconstruct the masked image regions based on these local features. As a result, the visual and textual global features output by the image and text encoders are already enriched with local information relevant to the current task. Compared to existing models that require statistical priors or off-the-shelf segmentation models for capturing local features of pedestrian images, our TG-MIM needs no priors or annotations and can effectively enhances the model's ability to understand detailed information within the query and align them with local visual cues in pedestrian images. Secondly, to further alleviate the person identity confusion issue, we introduce an Identity Supervised Global Visual Feature Calibration (IS-GVFC) task. This task utilizes the inherent pedestrian ID annotations present in the training samples and constructs visual feature distribution matching loss to minimize the distribution difference between images with same identity while maximize the difference between images having different identities. This helps the image encoder to possess a stronger ability to resist person identity confusion. In addition, different from the traditional global-local alignment methods, which need to align the local features of image and text in the inference stage, our model only needs to extract the global features of image and text and calculate their similarity in the inference stage.

Our contribution can be summarized as follows:

\begin{itemize}

\item We propose a Text Guided Masked Image Modeling (TG-MIM) auxiliary task, predicting occluded regions through text query and image context information. Our research indicates that this auxiliary task, which does not require additional supervision, can effectively enhance the image encoder's ability of understanding image content.

\item We design an Identity Supervised Global Visual Feature Calibration (IS-GVFC) auxiliary task. Through the visual feature distribution matching loss, IS-GVFC can guide the image encoder to focus on learning visual features related to person identities, thus improving the model's resistance to identity confusion problem.

\item Our research indicates that despite the significant differences between pedestrian images and natural scene images, the knowledge embedded in the pre-trained CLIP model can be successfully transferred and adapted to text-based person search task under the guidance of our designed auxiliary tasks.

\item Extensive experimental results on benchmarks illustrate that our model consistently outperforms the state-of-the-arts by a significant margin.
\end{itemize}

The remaining of this paper is arranged as follows: In section \ref{sec:relatedWorks}, we review and discuss the recent research on text-based person search task. We then provide a detailed introduction to our model in section \ref{sec:methodology}, focusing on the specific design and training methods of the two auxiliary tasks. In section \ref{sec:experiments}, we conduct comprehensive comparative experiments and ablation studies on three benchmarks to analyze the advancement of our model. Finally, we conclude our work in section \ref{sec:conclusion}.

\section{Related works}
\label{sec:relatedWorks}

\subsection{Global alignment approaches}
Text-based person search task was initially proposed by \emph{Li} et al. \cite{R1}, who also released the first benchmark CUHK-PEDES. This task requires the model to understand the natural language query input by the user and retrieve target pedestrians from a large-scale image database, and has received wide attention from both academia and industry in recent years due to its clear and promising application prospects. Early approaches extract global features from pedestrian image and text query, project them into a common latent space to achieve cross-modal alignment. For instance, in \cite{R1,R22}, VGG \cite{R23} and LSTM \cite{R24} were adopted to extract global visual and text features and attention mechanism was introduced to align these features. However, these methods are hard to capture the local details in images, such as the color of a person's clothing, the type of pants, and other attributes, which are crucial for the retrieval task.

\subsection{Global-local alignment approaches}
Subsequent research has shifted its focus to extracting local features from pedestrian images and text queries, aligning global and local aspects of images and texts. The challenge lies in accurately extracting local features from images. The methods for extracting local details from pedestrian images can roughly be divided into three categories: (1) The approaches involving segmenting pedestrian images using off-the-shelf segmentation models. Then local features are extracted from the segmented regions. For instance, ViTAA model \cite{R12} utilized the Human Parsing Network \cite{R15} to segment pedestrian image into regions and then aligned these regions with text phrases one by one. On the one hand, the limited generalization of the segmentation model may lead to errors in segmenting the pedestrian body. On the other hand, the introduction of the segmentation model increases the inference time. (2) The approaches dividing the pedestrian image directly into evenly sized patches along the vertical direction. In the literature \cite{R6,R10,R12,R25,R26,R39,R40,R47}, the feature maps extracted by convolutional neural networks were vertically segmented into six equally-sized patches, and pooling operations were then performed on the feature maps of each region to obtain local features of the pedestrian image. This kind of segmentation method can roughly segment pedestrians into head, upper body, waist, lower body, and feet without complex calculations. However, the drawback is that such segmentation method leads to inaccurate extraction of local features. (3) The approaches uniformly dividing pedestrian image into patches of size $P\times P$ and using an image encoder based on Transformer \cite{R35} to extract local features from each patch. For example, \emph{Han} et al. \cite{R18} firstly introduced CLIP into text-based person search task. Literature \cite{R27} proposed a simple framework for person search driven by pre-trained CLIP ViT and Bert \cite{R37} to extract local and global features from both images and texts separately, and aligning cross-modal information at both global and local scales. Recent works \cite{R28,R29,R30} have all adopted similar approaches. Due to the multi-head self-attention mechanism used in each encoding layer of ViT, there can be sufficient interaction between image regions, thus effectively alleviating the issue of accurate local feature extraction caused by uniform image segmentation. Moreover, this approach partially transfers the rich cross-modal knowledge embedded in pre-trained CLIP to text-based person search task, leading to significant improvements in retrieval accuracy. However, these approaches still have several disadvantages: (1) Complicated cross-modal local alignment is still needed. For example, CFine \cite{R19} designed cross-grained and fine-grained local alignment. The method proposed in this study discards the complex local alignment and instead guides the model to learn local information and integrate it into global features through cross-modal  masked image modeling auxiliary training. (2) The existing CLIP based models generally cannot address the issue of person identity confusion problem and their ability of obtaining discriminative visual features is still need to be enhanced. In this study, we also utilize the image encoder and text encoder from pre-trained CLIP to extract basic visual and text features. Building upon this, we design identity supervised global visual feature calibration task. Our work demonstrates that by simply incorporating appropriate auxiliary tasks during training, we can effectively transfer CLIP's knowledge to solve text-based person search task without the need for intricate local alignment when inference.

\section{Methodology}
\label{sec:methodology}

\begin{figure*}
\centering
\includegraphics[width=\linewidth]{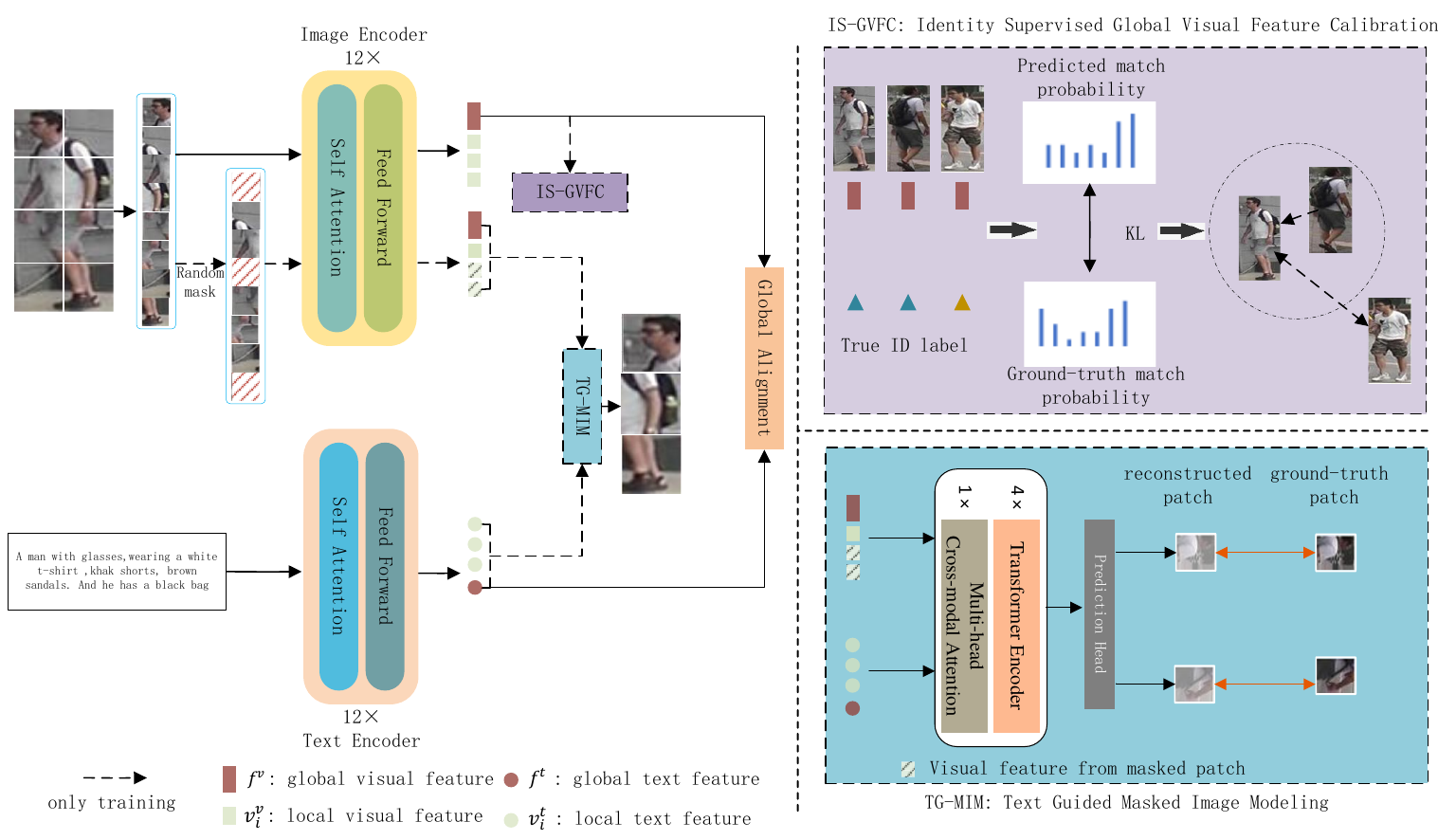}
\caption{The overall framework of our Visual Feature Enhanced Text-based Person Search (VFE-TPS) model. The model consists of a basic feature extraction module composed of an image encoder and a text encoder, a visual feature enhancement module (including Text Guided Masked Image Modeling (TG-MIM) shown in the blue dashed box and Identity Supervised Global Visual Feature Calibration (IS-GVFC) shown in the purple dashed box), and a cross-modal global alignment module. Dashed arrows and dashed boxes indicate that these paths or modules are only active during the training stage.}
\label{fig:modelOverview}
\end{figure*}

We propose a Visual Feature Enhanced Text-based Person Search model (VFE-TPS) as shown in Figure \ref{fig:modelOverview}. The model mainly consists of the following three parts: (1) Basic feature extraction module including image encoder and text encoder, which are utilized to learn the fundamental global and local representations of input images and text queries. (2) Visual feature enhancement module which conducts auxiliary tasks including Text Guided Masked Image Modeling (TG-MIM) task and Identity Supervised Global Visual Feature Calibration (IS-GVFC) task. The detail of IS-GVFC is demonstrated in the purple box while the blue box shows our TG-MIM. These auxiliary tasks are only executed during the training phase to enhance the image encoder's ability to learn task-relevant visual features with no additional annotations. (3) Cross-modal global alignment, which is used to calculate similarity based on the global representations of images and texts. This alignment enables the model to return target pedestrians that match the query description.

\subsection{Basic Feature Extraction}
\label{subsec:basicFeatureExtract}
Recent works \cite{R27,R28,R29,R30} have shown that, utilizing the vision-language models such as CLIP pre-trained on multimodal data can provide foundational features for cross-modal retrieval tasks, as well as offer essential cross-modal alignment capabilities. Therefore, in this study, we directly initialize our image encoder and text encoder using the pre-trained CLIP model.

\textbf{Text Encoder}. For the text query $T$, we directly employ the CLIP-Xformer \cite{R17} as the text encoder to extract global and local foundational features from the query. Specifically, we first truncate or pad the query to a sequence with fixed length $M$ and encode it using lower-case Byte Pair Encoding (BPE) \cite{R36}. Additionally, special tokens [SOS] and [EOS] are inserted to the beginning and end of the encoded sequence. Then position embedding is also introduced to generate the token sequence $\{token_{SOS}^t, token_{1}^t..., token_{M}^t, token_{EOS}^t\}$, which is further fed into the text encoder composed of 12 layers of multi-head self-attention modules to explore interaction between words. Suppose the output of the final layer is $\{v_{SOS}^t, v_1^t, ..., v_M^t, v_{EOS}^t\}$. $v_i^t$ which corresponds to the $i^{th}$ word of the query can be viewed as local text representation. $v_{EOS}^t$, the output tensor corresponding to the [EOS] token, thoroughly integrates information from all the words in the query. And this tensor is linearly projected through a fully connected layer to generate global text representation  $f^t=FC(v_{EOS}^t)$.

\textbf{Image Encoder}. For each pedestrian image $I\in \mathbb{R}^{C\times H\times W}$ where $C$, $H$, and $W$ denote the channel count, height and width of the image respectively, we utilize CLIP ViT as the image encoder to learn basic visual representation. Firstly, the image is uniformly divided into $N=H\times W/P^2$ patches with fixed size $P$ and then $Conv2d$ layer is adopted to linearly project these patches into token sequence $\{token_{CLS}^v, token_{1}^v..., token_{N}^v\}$, where $token_{CLS}^v$ is an inserted special token. After adding learnable position embeddings, CLIP ViT encodes this token sequence into $\{v_{CLS}^v, v_1^v, ..., v_N^v\}$, where $v_i^v$ is the representation of the $i^{th}$ image patch. And $v_{CLS}^v$ is further linearly projected through a fully connected layer to generate global visual representation  $f^v=FC(v_{CLS}^v)$.

\subsection{Visual Feature Enhancement}
\label{subsec:VFE}
The extracted visual features mentioned above have already captured the fundamental characteristics of pedestrians, enabling the assessment of the matching degree between each pedestrian and the query. Current mainstream models only utilize coarse constraint information such as image-text pair to supervise the training of the image encoder. Thus, the encoder lacks a strong understanding of visual details in the images and cannot deeply explore subtle information that is required by the query. In order to fully explore task-relevant visual details in pedestrian images, this study proposes Text Guided Masked Image Modeling and Identity Supervised Global Visual Feature Calibration. These auxiliary tasks are only conducted during the training phase to enhance the encoder's visual content understanding ability.

\begin{figure*}
\centering
	\subfloat[]{\includegraphics[width = 0.4\linewidth]{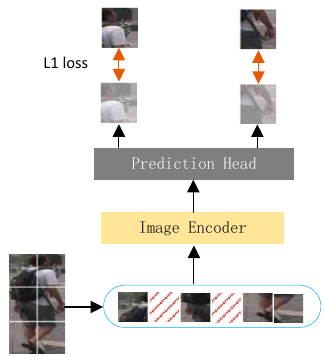}}
	\hspace{20mm}
	\subfloat[]{\includegraphics[width = 0.65\linewidth]{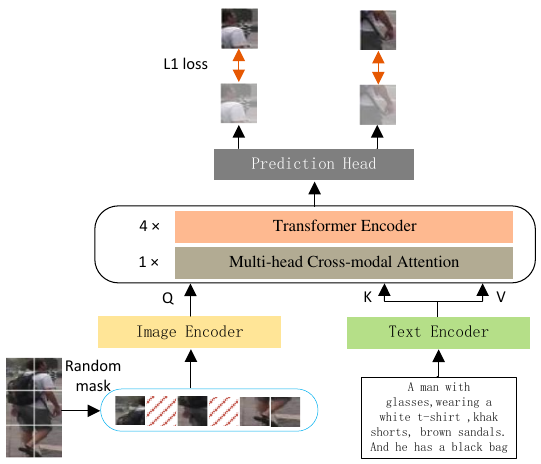}}
\caption{Illustration of our TG-MIM and the popular SimMIM. (a) SimMIM predicts raw pixel values for masked patches based on image context. (b) Our proposed TG-MIM firstly conduct cross-modal interaction via multi-head cross attention layer, and then predict raw pixel values for masked patches based on text and image context.}
\label{fig:MIM}
\end{figure*}

\subsubsection{Text Guided Masked Image Modeling}
\label{subsubsec:TGMIM}
Numerous works \cite{R6,R10,R12,R25,R26} have demonstrated that extracting detailed features related to the current retrieval task is crucial and challenging for the success of text-based person search task. Current approaches only utilize pre-trained CLIP to extract local visual features, lacking cues from query texts. As a result, their capability to learn task-relevant visual details is insufficient \cite{R28,R29}. Recently, inspired by the masked language modeling task used to train large language models, similar task named SimMIM \cite{R21} is proposed in the field of computer vision. As shown in Figure \ref{fig:MIM}(a), SimMIM reconstructs image patches based on the image context, thereby effectively enhancing image encoder's capability of visual understanding. This learning paradigm requires no extra annotations and fully leverages the intrinsic constraints within the images. The trained image encoder can be widely applied to downstream tasks such as image classification, semantic segmentation, and object detection. However, SimMIM lacks cross-modal capability and predicts masked regions only according to image context. In order to enable a text-based person search model to explore the local subtle information in pedestrian images based on query, we design Text Guided Masked Image Modeling (TG-MIM) auxiliary task which is shown in Figure \ref{fig:MIM}(b)). This task conducts cross-modal interaction to fully exploit the associated information between query and image regions, and then predicts raw pixel values for the masked regions based on cross-modal information. Thereby, the visual understanding capability of the image encoder can be enhanced. As shown in Algorithm \ref{alg:tgmim_code}, TG-MIM mainly consists of the following three steps:

\begin{algorithm}
\caption{ Computational flow of TG-MIM.}
\label{alg:tgmim_code}
\LinesNumbered
\KwIn{$\mathcal{B}$: A mini-batch of training query-image pairs. }

\ForEach{image $I$ and query $T$ pair in $\mathcal{B}$}{
    Uniformly dividing image $I$ into $N$ patches.\\
    Randomly masking image patches.\\
    Extracting visual features using image encoder: $z^v=\{f^v, v_1^v, ..., v_N^v\}=ImageEncoder(I)$.\\
    Extracting text features using text encoder: $z^t=\{f^t, v_1^t, ..., v_M^t\}=TextEncoder(T)$.\\
    Cross-modal interaction using MCA layers:\\
    $\tilde{z^v}=MCA(z^v, z^t, z^t)$\\
    Predicting pixels in each masked patch:\\
    $Y_P=PixelShuffle(Conv2d(\tilde{v_i^v}))$\\
    Calculating loss function: $\mathcal{L}_{TD-MIM}=\frac{1}{P^2}|Y_P-X_P|_1$\\
}
\end{algorithm}

\textbf{Randomly masking image regions}. Following \cite{R21}, patch-level random masking strategy is adopted, which means a certain proportion of patches are randomly masked. Since image patch is the basic processing unit for the image encoder composed of multi-head self-attention layers, patch-level random masking can be easily implemented by randomly selecting a number of patches and setting pixel values within these patches to zero.

\textbf{Features extraction}. We use CLIP ViT and Xformer to extract visual and text features from pedestrian image and query respectively. Since it has been introduced in section \ref{subsec:basicFeatureExtract}, we will not reiterated here. Let's denote visual feature as $z^v=\{f^v, v_1^v, ..., v_N^v\}$ where $f^v$ and $v_i^v$ represents the global and local visual features respectively. Similar, we use $z^t=\{f^t, v_1^t, ..., v_M^t\}$ to denote features of the input query.

\textbf{Cross-modal interaction and pixel prediction}. It is an important distinction between our TG-MIM and the popular SimMIM. Firstly, we utilize the visual feature $z^v$ as the query, and the text feature $z^t$ as the key and value. Then cross-modal interaction can be conducted through one Multi-head Cross-modal Attention (MCA) layer, which integrates the semantic information from the query to enhance the relevant details in the image for retrieval tasks. The calculation method is as follows:
\begin{equation}
\label{eq:mca}
\tilde{z^v}=MCA(z^v, z^t, z^t) = [head_1, ..., head_H]\cdot W^O
\end{equation}
\begin{equation}
\label{eq:mcaHead}
head_i=softmax(\frac{(z^vW_i^Q)\cdot(z^tW_i^K)^T}{\sqrt{d}})z^tW_i^V
\end{equation}
where $H$ denotes the heads of cross attention. $W_i^Q, W_i^K, W_i^V\in \mathbb{R}^{d\times d_h}$ are learnable matrix in the $i^{th}$ attention head. $W^O\in \mathbb{R}^{H*d_h\times d}$ is the output projection matrix. $d_h=d/H$ means the model dimension of each head. The output of MCA module is further explored and fused through four Transformer encoder layers, resulting in an enhanced image representation. This enhanced representation is then fed into a prediction head composed of a single \emph{Conv2D} layer to predict the raw pixel values and reconstruct the masked regions using $PixelShuffle$ \cite{n4} operation. Similar to SimMIM, our TG-MIM does not require extra annotation and can be trained under the supervision of $L1$ loss as below,
\begin{equation}
\label{eq:l1loss}
\mathcal{L}_{TD-MIM}=\frac{1}{P^2}|Y_P-X_P|_1
\end{equation}
where $P$ denote the patch size, $X_P$ and $Y_P$ respectively denote the predicted and ground-truth pixel values of the masked patch.

\textbf{Discussion}. \textit{"The devil lies in the details"}. Extracting local detailed features from both query text and pedestrian images is crucial for improving the accuracy of person search. Existing methods generally rely on pre-trained visual models or prior knowledge, along with text parsers, to obtain local features of images and text. They also need design complex cross-modal local alignment modules. However, our designed TG-MIM ingeniously leverages image context correlation, utilizing image context information to predict the content of masked regions, thereby enabling the encoder to possess the ability to mine local details of images. Furthermore, our TG-MIM differs from traditional MIM in that it introduces semantic information from the query text as guidance when predicting masked regions, thereby enhancing its cross-modal capability and enabling the model to achieve implicit cross-modal alignment.

\begin{figure}
\centering
\includegraphics[width=0.6\linewidth]{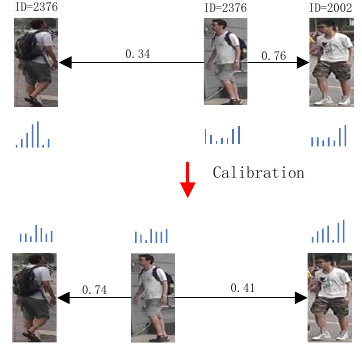}
\caption{Illustration of our Identity Supervised Global Visual Feature Calibration (IS-GVFC). The number on the arrow line denotes the similarity between images. IS-GVFC can reduce the difference between global visual features of pedestrian images with same identity, while expand the distance between pedestrian images with different identities.}
\label{fig:ISGVFC}
\end{figure}

\subsubsection{Identity Supervised Global Visual Feature Calibration}
\label{subsubsec:ISGVFC}
In text-based person search task, assuming the user describes pedestrian $A$, then all images of this pedestrian captured by different cameras are considered as target images. Therefore, the model should assign high scores to the images with same identity even if their apparent features are quite different. Hence, mitigating identity confusion (as shown in Figure \ref{fig:motivation}(a)) is the key for improving the retrieval accuracy. In this study, we propose an Identity Supervised Global Visual Feature Calibration (IS-GVFC) auxiliary task to guide the model to learn robust global visual features. As shown in Figure \ref{fig:ISGVFC}, IS-GVFC can guide the image encoder to learn ID-aware visual features, so as to narrow the distribution divergence of pedestrians with same ID, and increase the visual distance between pedestrians with different IDs.

\begin{algorithm}
\caption{ Computational flow of IS-GVFC.}
\label{alg:isgvfc_code}
\LinesNumbered
\KwIn{$\mathcal{B}$: A mini-batch of training query-image pairs. }
Randomly sampling $C$ image pairs $(I_i, I_j)$ from $\mathcal{B}$.\\
Getting images' global visual feature for each image: $f_i^v$ and $f_j^v$.\\
Annotating each sample: $y_{i,j}=1$ if image $I_i$ and $I_j$ share the same identity.\\
Predicting matching probability between images using equation \ref{eq:matchingP}\\
Calculating KL distance between predicted matching probability and ground-truth probability using equation \ref{eq:klLoss}.\\
\end{algorithm}

Similar to TG-MIM, IS-GVFC is performed only in the training stage. Specifically, a sample set $\{(f_1^v, pid(f_1^v)), ..., (f_B^v, pid(f_B^v))\}$ ($B$ denotes the batch size) is collected in each training iteration, where $f_i^v$ is the global visual feature of image $i$ and $pid(f_i^v)$ is its corresponding person identity. Then a set of image pairs are sampled, denoted as $\{(f_i^v, f_j^v), y_{i,j}\}_{i,j=1}^C$. $y_{i,j}=1$ if image $i$ and image $j$ have same person identity, otherwise $y_{i,j}=0$. Since the purpose of IS-GVFC is to guide the model to learn identity-aware visual representations, the following KL distance between visual feature matching probability is adopted,
\begin{equation}
\label{eq:klLoss}
\mathcal{L}_{IS-GVFC}=\frac{1}{C}\sum_{i=1}^C\sum_{j=1}^CKL(p_{i,j}^{v,v}||q_{i,j}^{v,v})
\end{equation}
where $q_{i,j}^{v,v}=\frac{y_{i,j}}{\sum_{k=1}^Cy_{i,k}}$ denotes the ground-truth matching probability. $p_{i,j}^{v,v}$ is the predicted matching probability based on global visual features,
\begin{equation}
\label{eq:matchingP}
p_{i,j}^{v,v}=\frac{exp(cos(f_i^v,f_j^v)/\tau)}{\sum_{k=1}^Cexp(cos(f_i^v, f_k^v)/\tau)}
\end{equation}
where $\tau$ is a temperature hyperparameter used to control the probability distribution peaks. Minimizing this KL divergence can guide the image encoder to learn identity-aware visual features. Algorithm \ref{alg:isgvfc_code} demonstrates the detailed computational flow of our proposed IS-GVFC.

\textbf{Discussion}. Current approaches \cite{R6,R10,R25,R26,R39} typically introduce an additional pedestrian identity recognition head or pedestrian similarity ranking head into the model, and guide the model to learn identity-related features by incorporating ID loss or triplet loss. However, our preliminary research \cite{R6} indicated that such classification or ranking task is inadequate for effectively calibrating visual features and does not fully address the issue of identity confusion. Moreover, calculating triplet loss requires selecting triplets from the training batch (including an anchor image, a positive sample with the same ID as the anchor, and a negative sample with different ID) and tuning the margin parameter. In comparison, our proposed IS-GVFC is much simpler. IS-GVFC guides the model to learn identity-related features by minimizing the KL divergence between the distribution of image global feature similarity and the label matching distribution. We will quantitatively analyze the above three strategies in the experimental section \ref{subsubsec:ablation_ISGVFC}.

\subsection{Cross-modal Global Alignment}
\label{subsec:CMGA}

Unlike existing approaches that require aligning local features of images and texts one by one, our model only needs to compute the similarity between global representations during training and inferencing stages. In this study, Cross Model Projection Match (CMPM) loss \cite{R9} is adopted to supervise the global alignment. Let's use $f_i^v$ to denote the global visual representation for image $i$ and $f_j^t$ to be the global representation for query $j$. We construct positive and negative image-text pairs $\{(f_i^v, f_j^t), y_{i,j}\}_{j=1}^B$ for image $i$, where $y_{i,j}=1$ if image $i$ matches query $j$ otherwise $y_{i,j}=0$. Then we can predict the matching probability of image $i$ and query $j$ as follows,
\begin{equation}
\label{eq:P}
p_{i,j}^{v,t}=\frac{exp(cos(f_i^v,f_j^t)/\tau)}{\sum_{k=1}^Bexp(cos(f_i^v, f_k^t)/\tau)}
\end{equation}
Thus, image-to-text matching loss for image $i$ can be calculated as follows,
\begin{equation}
\label{eq:Li}
\mathcal{L}_i = \frac{1}{B}\sum_{j=1}^Bp_{i,j}^{v,t}log(\frac{p_{i,j}^{v,t}}{q_{i,j}^{v,t}+\epsilon})
\end{equation}
where $\epsilon$ is a small constant used to avoid numerical problems. $q_{i,j}^{v,t}$ is the ground-truth matching probability which can be obtained as $q_{i,j}^{v,t}=y_{i,j}/\sum_{k=1}^By_{i,k}$. By utilizing equation \ref{eq:Li}, the CMPM loss in image-to-text direction is the average of all image-to-text matching losses, which can be obtained as follows,
\begin{equation}
\label{eq:LI2T}
\mathcal{L}_{I2T}=\frac{1}{B}\sum_{i=1}^B\mathcal{L}_i
\end{equation}
The CMPM loss in text-to-image direction $\mathcal{L}_{T2I}$ can be calculated in the same way. The final CMPM loss is as follows,
\begin{equation}
\label{eq:CMPM}
\mathcal{L}_{CMPM}=\mathcal{L}_{I2T}+\mathcal{L}_{T2I}
\end{equation}

In addition, combining the TG-MIM and IS-GVFC auxiliary tasks proposed in Section \ref{subsec:VFE}, our propopsed VFE-TPS model can achieve end-to-end training under the supervision of the following overall loss function,
\begin{equation}
\label{eq:finalLoss}
\mathcal{L}=\mathcal{L}_{TG-MIM}+\mathcal{L}_{IS-GVFC}+\mathcal{L}_{CMPM}
\end{equation}

\section{Experiments}
\label{sec:experiments}

In this section, extensive experiments are conducted on three challenging text-based person search datasets to evaluate our proposed approaches.

\subsection{Datasets}
\label{subsec:dataset}

\textbf{CUHK-PEDES} \cite{R1} is the first dataset for text-based person search task. It consists of 13,003 identities captured by surveillance systems from various angles. Each image is associated with two descriptive texts, totaling 40,206 images and 80,422 texts. We follow the partitioning method in \cite{R1} to divide the dataset into training, validation, and test sets. The training set includes 34,054 images and 68,108 texts from 11,003 individual identities. The remaining image-text pairs are evenly split to generate the validation and test sets.

\textbf{ICFG-PEDES} \cite{R26} comprises 54,522 images of 4,102 identities captured in various settings. Each pedestrian is represented by 5 images, and each image is paired with one descriptive texts. Following the official data split, we can obtain the training set composed of 34,674 image-text pairs from 3,102 identities, and the test set composed of 19,848 image-text pairs from 1,000 identities.

\textbf{RSTPReid} \cite{R39} is a recently released dataset. It contains 20,505 images from 4,101 identities. Each image has 2 text descriptions. Following the official data split, the training, validation and test set respectively contain 3,701, 200, and 200 identities.

\subsection{Evaluation Metrics}
\label{subsec:evaMetric}
In order to facilitate a fair comparison with existing research, we adopt the evaluation metrics in the field of cross-modal retrieval, including Rank-$k$($k$=1,5,10) and mean Average Precision (mAP). The Rank-$k$ metric involves sorting the searching results in descending order and calculating the probability that at least one hit is found within the top $k$ results. Meanwhile, mAP provides more comprehensive assessment of algorithm performance. The higher Rank-$k$ and mAP indicate better performance.

\subsection{Implementation Details}
\label{subsec:impDetails}
Due to the inconsistent sizes of pedestrian images, we first uniformly resize the images to $384\times128$. Additionally, data augmentation such as random horizontal flip and random crop are applied. The input query text is standardized to fixed length 77 ($M=77$) by truncating or zero-padding. We initialize our image encoder and text encoder using the pre-trained CLIP-ViT-B/16 and CLIP-Xformer respectively. Consequently, each image is evenly divided into 192 ($N=192$) patches. We set $H=8$ and $d=768$ for the MCA layers in TG-MIM. Adam optimizer is adopted to train our model for 60 epoches. The initial learning rate is set to $1\times 10^{-5}$ and batch size $B$ is set to 100. To calculate $\mathcal{L}_{IS-GVFC}$ when training, we randomly sample 20 ($C=20$) image pairs.  We implement our model using PyTorch and our experiments are all conducted on a computer with a single Nvidia RTX 3090 GPU and Intel Core i9 CPU.

\renewcommand{\arraystretch}{1}
\begin{sidewaystable}
  \centering
  \begin{threeparttable}
  \caption{Overall comparison between our model and SOTAs on CUHK-PEDES.}
  \label{tab:SOTA_compair_CUHK}
    \begin{tabular}{@{}ccccccccccccccc@{}}
    \toprule
    Model & Reference &   Image enc  & Text Enc & Rank-$1$ & Rank-$5$ & Rank-$10$ & mAP \cr
    \midrule
    TIMAM\cite{R47} & ICCV 2019 & RN101 & Bert & 54.51 & 77.56 & 79.27 & - \\
    MIA\cite{R25} &	TIP 2020 & RN50 & GRU & 53.10 & 75.00 & 82.90 & - \\
    ViTAA\cite{R12} & ECCV 2020 &RN50 &LSTM &54.92 &75.18 &82.90 &51.60 \\
    NAFS\cite{R40} &arXiv 2021&RN50 &BERT &59.36 &79.13 &86.00 &54.07 \\
    DSSL\cite{R39} &ACM MM 2021 &RN50 &BERT &59.98 &80.41 &87.56 &- \\
    TPSLD\cite{R18} &BMVC 2021 &CLIP-RN101 &CLIP-Xformer &64.08 &81.73 &88.19 &60.08 \\
    SRCF\cite{R41} &ECCV 2022  &RN50 &BERT &64.04 &82.99 &88.81 &- \\
    SAF\cite{R42} &ICASSP 2022  &ViT &BERT &64.13 &82.62 &88.4 &58.61 \\
    AXM-Net\cite{R43} &AAAI 2022 &RN50 &BERT &64.44 &80.52 &86.77 &58.73 \\
    LGUR\cite{R49} &ACM MM 2022 &DeiT &BERT &64.21 &81.94 &87.93 &- \\
    IVT\cite{R44} &ECCV 2022 &ViT &BERT &65.59 &83.11 &89.21 &- \\
    CSKT\cite{n1} & arXiv 2023 & CLIP-ViT & CLIP-Xformer & 69.70 & 86.92 & 91.8 & 62.74 \\
    PLIP\cite{R25} &arXiv 2023 &RN50 &GRU &68.16 &85.56 &91.21 &- \\
    CFine\cite{R19} &TIP 2023 &CLIP-ViT &BERT &69.57 &85.93 &91.15 &- \\
    CM-LRGNet\cite{R6} &KBS 2023 &RN50 &BERT &64.18 &82.97 &89.85 &- \\
    TP-TPS\cite{R51} &arXiv 2023 &CLIP-ViT &CLIP-Xformer &70.16 &86.10 &90.98 &66.32 \\
    MGEL\cite{R13} &TMM 2024 &RN50 &BiLSTM &60.27 &80.01 &86.74 &- \\
    VGSG\cite{R46} &TIP 2024 &CLIP-ViT &CLIP-Xformer &71.38 &86.75 &91.86 &- \\
    \hline
    baseline (ours) &  &ViT &Bert &70.12 &87.96 &92.78 &63.69\\
    VFE-TPS (ours) & &CLIP-ViT &CLIP-Xformer &\textbf{72.47} &\textbf{88.24} &\textbf{93.24} &\textbf{64.26}\\
    \bottomrule
    \end{tabular}
    \end{threeparttable}
\end{sidewaystable}

\subsection{Overall Comparison with State-Of-The-Arts}
\label{subsec:compSOTA}
To validate the superiority of our proposed algorithm, current State-Of-The-Art (SOTA) algorithms are introduced to compare with our approach. The experimental results are reported in Table 1 to Table 3, where image and text encoders used by the models, along with their retrieval performance, are reported. Since most recent SOTAs use ViT and Bert pre-trained on unimodal data, we also design a baseline model which shares the same architecture with VFE-TPS but utilizes the same encoders with these SOTAs.

\textbf{Results on CUHK-PEDES}. Table \ref{tab:SOTA_compair_CUHK} reports the performance of our model and the SOTAs. The Rank-$1$ accuracy of our VFE-TPS model reaches 72.47\%, the Rank-$5$ accuracy reaches 88.24\%, and the Rank-$10$ accuracy reaches 93.24\%, surpassing all the SOTAs. We observe that a major trend in text-based person search task is the adoption of more powerful image and text encoders (such as ViT and Bert), leading to significant improvements in retrieval accuracy. Compared to SOTA models such as CFine \cite{R19}, VGSG \cite{R46} and TP-TPS \cite{R51} that also utilize pre-trained CLIP as image and text encoders, our model has achieved an improvement of about 1\%$\sim$3\%, which is contributed by our designed auxiliary tasks including TG-MIM and IS-GVFC.

\renewcommand{\arraystretch}{1}
\begin{sidewaystable}
  \centering
  \begin{threeparttable}
  \caption{Overall comparison between our model and SOTAs on ICFG-PEDES.}
  \label{tab:SOTA_compair_ICFG}
    \begin{tabular}{@{}ccccccccccccccc@{}}
    \toprule
    Model & Reference &   Image Enc  & Text Enc & Rank-$1$ & Rank-$5$ & Rank-$10$ & mAP \cr
    \midrule
    MIA\cite{R25} &TIP 2020 &RN50 &GRU &46.49 &67.14 &75.18 &-\\
    ViTAA\cite{R12} &ECCV 2020 &RN50 &LSTM &50.98 &68.79 &75.78 &-\\
    SSAN\cite{R26} &arXiv 2021 &RN50 &LSTM &54.23 &72.63 &79.53 &-\\
    IVT\cite{R44} &ECCV 2022 &ViT &BERT &56.04 &73.60 &80.22 &-\\
    SRCF\cite{R41} &ECCV 2022 &RN50 &BERT &57.18 &75.01 &81.49 &-\\
    LGUR\cite{R49} &ACM MM 2022 &DeiT &BERT &59.02 &75.32 &81.56 &-\\
    MANet\cite{R48} &arXiv 2023 &RN50 &BERT &59.44 &76.80 &82.75 &-\\
    CFine\cite{R19} &TIP 2023 &CLIP-ViT &BERT &60.83 &76.55 &82.42 &-\\
    CSKT\cite{n1} & arXiv 2023 & CLIP-ViT & CLIP-Xformer & 58.90 & 77.31 & 83.56 & 33.87 \\
    TP-TPS\cite{R51} &arXiv 2023 &CLIP-ViT &CLIP-Xformer &60.64 &75.97 &81.76 &42.78\\
    \hline
    baseline (ours) &  &ViT &Bert &60.98 &77.65 &83.51 &40.01\\
    VFE-TPS (ours) & &CLIP-ViT &CLIP-Xformer &\textbf{62.71} &\textbf{78.73} &\textbf{84.51} &\textbf{43.08}\\
    \bottomrule
    \end{tabular}
    \end{threeparttable}
\end{sidewaystable}

\textbf{Results on ICFG-PEDES}. The experimental results on ICFG-PEDES are shown in Table \ref{tab:SOTA_compair_ICFG}, which demonstrates that (1) A powerful feature extraction backbone significantly contributes to the retrieval accuracy promotion. For instance, CFine \cite{R19}, TP-TPS \cite{R51} and our proposed VFE-TPS utilize pre-trained CLIP to extract visual features, leading to significantly higher retrieval accuracy compared to traditional models like SSAN \cite{R26} and ViTAA \cite{R12} which use ResNet as image encoder. (2) Similar to the results on CUHK-PEDES dataset, the accuracy of our proposed model shows a significant improvement over current SOTA models. Compared to CFine, Rank-$1$, Rank-$5$ and Rank-$10$ of our VFE-TPS model have respectively increased by 1.88\%, 2.18\% and 2.09\%. Even our baseline model has outperformed all existing models. This indicates that our proposed TG-MIM and IS-VGFC can effectively enhance the image encoder's ability to understand visual details in the pedestrian images, assisting the model in achieving better retrieval performance.

\renewcommand{\arraystretch}{1}
\begin{sidewaystable}
  \centering
  \begin{threeparttable}
  \caption{Overall comparison between our model and SOTAs on RSTPReid.}
  \label{tab:SOTA_compair_RSTP}
    \begin{tabular}{@{}ccccccccccccccc@{}}
    \toprule
    Model & Reference &   Image Enc  & Text Enc & Rank-$1$ & Rank-$5$ & Rank-$10$ & mAP \cr
    \midrule
    DSSL\cite{R39} &ACM MM 2021 &RN50 &BERT &39.05 &62.60 &73.95 &-\\
    SSAN\cite{R26} &arXiv 2021 &RN50 &LSTM &43.50 &67.80 &77.15 &-\\
    LBUL\cite{R50} &ACM MM 2022 &RN50 &GRU &45.55 &68.20 &77.85 &-\\
    IVT\cite{R44} &ECCV 2022 &ViT &BERT &46.70 &70.00 &78.80 &-\\
    CFine\cite{R19} &TIP 2023 &CLIP-ViT &BERT &50.55 &72.50 &81.60 &-\\
    TP-TPS\cite{R51} &arXiv 2023 &CLIP-ViT &CLIP-Xformer &50.65 &72.45 &81.20 &43.11\\
    CSKT\cite{n1} & arXiv 2023 & CLIP-ViT & CLIP-Xformer & 57.75 & 81.30 & 88.35 & - \\
    \hline
    baseline (ours) &  &ViT &Bert &57.31 &79.98 &87.87 &44.01\\
    VFE-TPS (ours) & &CLIP-ViT &CLIP-Xformer &\textbf{59.25} &\textbf{81.90} &\textbf{88.85} &\textbf{45.96}\\
    \bottomrule
    \end{tabular}
    \end{threeparttable}
\end{sidewaystable}

\textbf{Results on RSTPReid}. As shown in Table \ref{tab:SOTA_compair_RSTP}, on the RSTPReid dataset, the advantages of our proposed model are even more pronounced. Our baseline model has already surpassed the current SOTA methods. Compared to CFine and TP-TPS models using the same image and text encoders, our VFE-TPS model has made significant progress, achieving breakthroughs with Rank-$1$, Rank-$5$, and Rank-$10$ of 59.25\%, 81.9\%, and 88.85\% respectively.

In summary, the approach proposed in this study has comprehensively outperformed current mainstream methods on the three standard datasets, demonstrating the advantages and generalization of our method. By comparing with other models utilizing CLIP as backbones, it is proved that the knowledge embedded in the pre-trained CLIP model can be successfully transferred and adapted to text-based person search task under the guidance of our designed
auxiliary tasks. For example, CFine \cite{R19} utilizes CLIP for text-based person search, which required intricate alignment of both global and local features during training and inference. Similarly, the VGSG \cite{R46} model, also built upon CLIP, performs complex cross-modal interactions and alignments across both global and local feature branches. In contrast, our model is concise, elegant, and efficient. Our work demonstrates that by simply incorporating appropriate auxiliary tasks during the training phase, without the need for intricate adjustments or additional modules, we can effectively transfer CLIP's knowledge to text-based person search tasks.

\renewcommand{\arraystretch}{1}
\begin{table*}[tp]
  \centering
  \begin{threeparttable}
  \caption{Computational complex comparison among state-of-the-art models on the CUHK-PEDES dataset. "Params" indicates the number of model parameters. "GPU mem" is the required GPU memory when training. "Inference" represents the inference time cost. }
  \label{tab:complex}
    \begin{tabular}{@{}ccccccccccccccc@{}}
    \toprule
    Model               &  Params(M)   & GPU mem(GB) & Inference(ms)  & Rank-1 \cr
    \midrule
    NAFS\cite{R40}       & 189    & 8.2 & 268 & 59.36 \\
    CM-LRGNet\cite{R6}   & 103    & 3.8 & 105 & 64.18 \\

    IVT\cite{R44}        & 166    & 7.5 & 660 & 65.59 \\
    CFine\cite{R19}      & 204    & 7.3 & 610 & 69.57 \\
    RaSa\cite{R29}            & 420    & 13.7& 1025& 76.51 \\
    \hline
    VFE-TPS(ours)        & 149    & 4.3 & 112 & 72.47\\
    \bottomrule
    \end{tabular}
    \end{threeparttable}
\end{table*}

\textbf{Computational Complexity Analysis}. We have also undertaken a series of experiments to assess the computational complexity of several SOTA models which have released their code, employing a consistent and equitable experimental setup. Specifically, when training, the batch size is set to 8, so that we can train these models in a single RTX 3090 GPU. And a batch size of 1 is set when inferencing. We report the number of model parameters (Params), GPU memory used when training (GPU mem), and inference time cost (Inference) in Table \ref{tab:complex}. The experimental results show that compared with the CFine and IVT, which are based on similar backbone, our model requires significantly fewer model parameters, GPU memory, and inference time, while achieving a notable improvement in retrieval accuracy. In addition, there are some recent works, such as RaSa \cite{R29}, have achieved higher retrieval accuracy. RaSa takes ALBEF \cite{albef} as its backbone, and achieves 76.51\% Rank-1 accuracies on CUHK-PEDES. However, training RaSa model requires a large amount of GPU memory. And its computational overhead is as high as 163 GFLOPs in the training and inference stages. In contrast, our model is much more efficient, which can be trained on a single RTX 3090 GPU with a computational overhead of only 17 GFLOPs during the training phase. And it can be further reduced to 12 GFLOPs in the inference phase, since the TG-MIM and IS-GVFC modules will be removed when inference.

\subsection{Ablation Study}
\label{subsec:ablation}
In this section, we design extensive ablation experiments to conduct in-depth analysis of each module proposed in this study. The following experiments are all conducted on the CUHK-PEDES dataset.

\subsubsection{Transferring CLIP to Text-based Person Search}
CLIP is pre-trained on 400M image-text pairs. Previous studies \cite{R17} have shown that CLIP has impressive zero-shot capability in a variety of computer vision downstream tasks. The natural question is: Does CLIP still have such zero-shot capability for text-based person retrieval task? To evaluate this idea, we conduct the following experiment. As demonstrated in Table \ref{tab:ablation_clip}, we initialize our model using pre-trained CLIP weights and then freeze these model weights, resulting in Zero-shot CLIP, which achieves poor retrieval accuracy. This experiment indicates that it is not feasible to directly use the pre-trained CLIP model to perform the text-based person search task. The reason may be that the CLIP model is trained using natural scene images and corresponding descriptions, which have significant difference from the pedestrian images and descriptions. However, the multimodal knowledge embedded in the pre-trained CLIP model can be transferred to our task. For example, the Linear probe CLIP also freezes CLIP weights but adds fully connected layer after the image encoder and text encoder respectively to linearly project the extracted feature. It is obvious that the Rank-1 accuracy rises rapidly from 12.61 to 24.16. Furthermore, after performing full parameter fine-tuning on CLIP using the CMPM loss, CLIP has achieved significant progress on the text-based person search task. Therefore, using CLIP as the backbone of a text-based person search model is a promissing choice, which is why most recently published SOTA models have adopted cross-modal pre-trained models such as CLIP as their backbones. In addition, the result of our VFE-TPS indicates that the knowledge embedded in the pre-trained CLIP model can be successfully transferred and adapted to text-based person search task under the guidance of our designed auxiliary tasks.

\renewcommand{\arraystretch}{1}
\begin{table*}[tp]
  \centering
  \begin{threeparttable}
  \caption{Performance comparison of zero-shot CLIP, linear probe CLIP and our model on CUHK-PEDES}
  \label{tab:ablation_clip}
    \begin{tabular}{ccccccccccccccc}
    \toprule
    Model  & Rank-$1$ & Rank-$5$ & Rank-$10$ & mAP \cr
    \midrule
    Zero-shot CLIP &     12.61 &  27.08   &  35.48   &  12.36\\
    Linear probe CLIP & 24.16 & 38.99 & 46.69 & 21.25 \\
    CLIP  & 66.78 & 85.29 & 91.20 & 59.68 \\
    VFE-TPS &\textbf{72.47} &\textbf{88.24} &\textbf{93.24} &\textbf{64.26}\\
    \bottomrule
    \end{tabular}
    \end{threeparttable}
\end{table*}

\subsubsection{Comprehensive Impact of Visual Feature Enhancement}
\label{subsubsec:impactVFE}

Compared to models like CFine \cite{R19} and TP-TPS \cite{R51} which also utilize CLIP, the significant innovation of our model lies in the design of visual feature enhancement auxiliary tasks including TG-MIM and IS-GVFC during training stage. These tasks enhance the image encoder's understanding of image details and improve the model's cross-modal alignment capability. To verify the impact of our visual feature enhancement auxiliary tasks, a comparative experiment is conducted, and the results are shown in Table \ref{tab:ablation_impact_VFE}. It is evident that the inclusion of auxiliary tasks during training stage leads to a corresponding improvement in retrieval accuracy. For instance, when incorporating IS-GVFC task alone, the Rank-$1$, Rank-$5$ and Rank-$10$ have been promoted by 1\%, 0.63\% and 0.53\% respectively. Moreover, the retrieval accuracy shows a more significant enhancement when introducing TG-MIM task. When both of these auxiliary tasks are added during training, the model's Rank-$1$, Rank-$5$ and Rank-$10$ can be significantly increased to 72.47\%, 88.24\%, and 93.24\%. The mAP metric has also promoted to 64.26\%.

\renewcommand{\arraystretch}{1}
\begin{table*}[tp]
  \centering
  \begin{threeparttable}
  \caption{Ablation study on each visual feature enhancement auxiliary task on CUHK-PEDES.}
  \label{tab:ablation_impact_VFE}
    \begin{tabular}{ccccccccccccccc}
    \toprule
    TG-MIM & IS-GVFC  & Rank-$1$ & Rank-$5$ & Rank-$10$ & mAP \cr
    \midrule
            &        &     70.61 &  87.66   &  92.32   &  63.12\\
    $\surd$ &        &     72.16 &  88.47   &  93.02   &  63.99\\
            &$\surd$ &     71.61 &  88.29   &  92.85   &  64.24\\
    $\surd$ &$\surd$ &     72.47 &  88.24   &  93.24   &  64.26\\
    \bottomrule
    \end{tabular}
    \end{threeparttable}
\end{table*}

\begin{figure}
\centering
\includegraphics[width=0.7\linewidth]{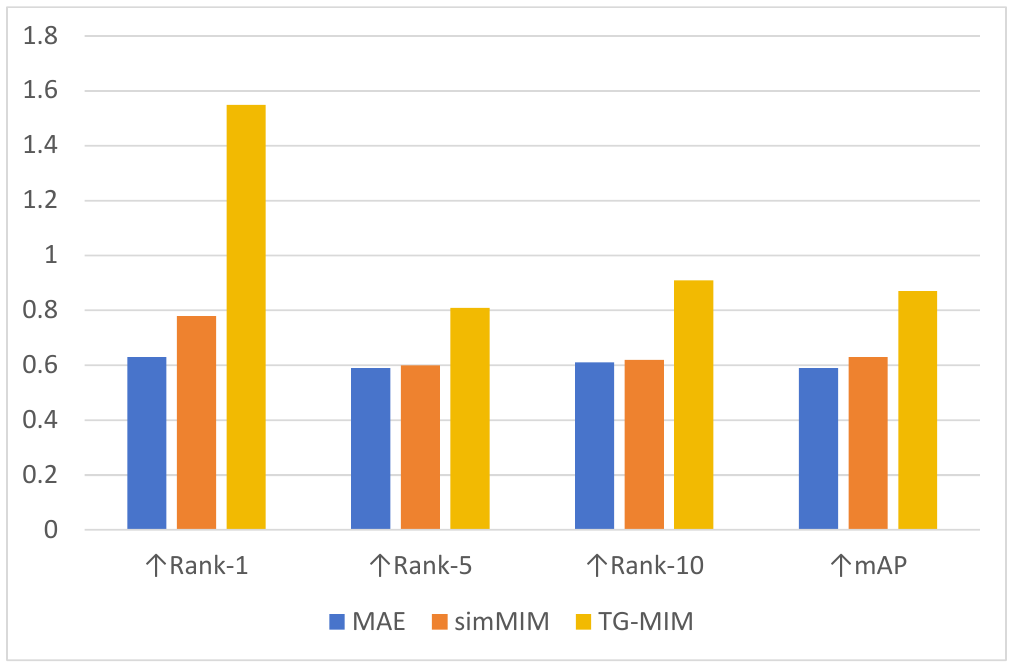}
\caption{Ablation study of different masked image modeling methods. $\uparrow$Rank-$k$ and $\uparrow$mAP denote the promotion of Rank-$k$ and mAP respectively.}
\label{fig:ablation_TGMIM}
\end{figure}

\subsubsection{Analysis of Text Guided Masked Image Modeling}
\label{subsubsec:ablation_MIM}
We argue that enhancing the visual understanding and modeling capability of image encoder is crucial for improving text-based person search accuracy. When an image encoder truly understands the content of an image, the model should possess the ability to generate the image content. Therefore, the Text Guided Masked Image Modeling presented in Figure \ref{fig:MIM}(b) is proposed. In order to validate the advancement of the proposed masked image modeling method, we compared it with two popular masking methods, namely MAE \cite{R20} and SimMIM \cite{R21}. We record the mAP of models trained using different masking methods and the experimental results are shown in Figure \ref{fig:ablation_TGMIM}. From the experimental results, it can be observed that introducing any masked image modeling auxiliary task during training phase leads to improvements. However, compared to strategies like MAE and SimMIM, which predict raw pixel values based solely on image context information, our proposed TG-MIM, integrating cross-modal information from both image context and query text, demonstrates more accurate understanding of pedestrian image content. Consequently, it achieves the most significant performance improvement.

\begin{figure}
\centering
	\subfloat[]{\includegraphics[width = 0.45\linewidth]{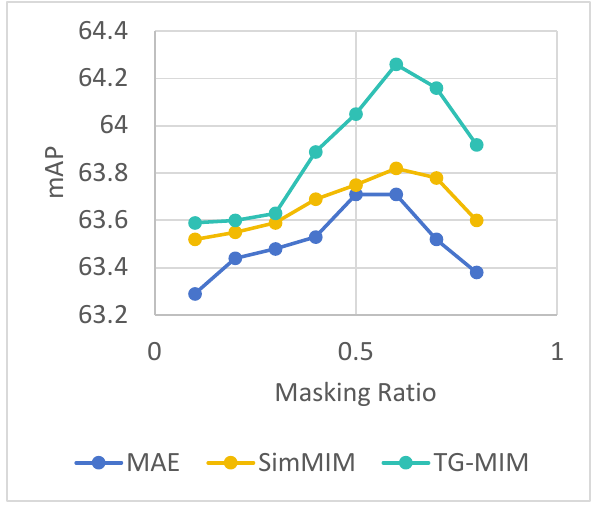}}
	\hspace{5mm}
	\subfloat[]{\includegraphics[width = 0.45\linewidth]{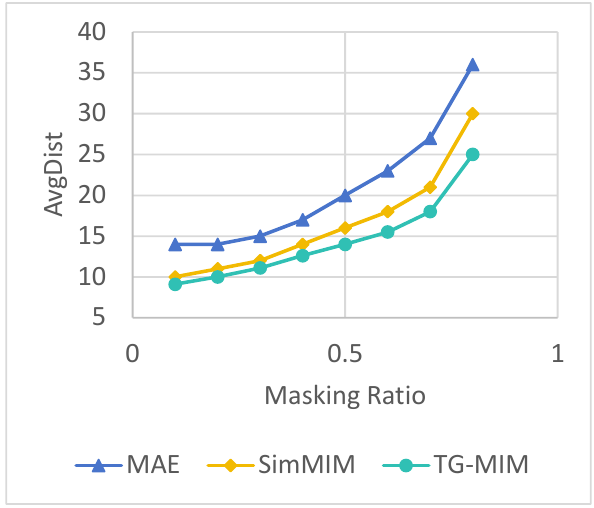}}
\caption{Ablation study of masked image modeling. (a) mAP curves of MAE, SimMIM and TG-MIM under different masking ratios. (b) Average distance of MAE, SimMIM and TG-MIM under different masking ratios.}
\label{fig:ablation_MIM}
\end{figure}

Additionally, the masking ratio is an important hyperparameter in masked image modeling task, and we also investigate the impact of this hyperparameter on the model. Figure \ref{fig:ablation_MIM}(a) shows the mAP curves of different masking methods at different masking ratios. From the experimental results, it can be seen that for the three masking methods, slight changes in mAP occur with different masking ratios. When the masking ratio is between 0.5 and 0.6, mAP reaches its peak. Furthermore, the performance of our proposed TG-MIM is better than MAE and SimMIM. To explore the reasons behind the above phenomenon, we further analyze the average distance (avgDist) \cite{R21} for different masking methods trained at different masking ratios, as shown in Figure \ref{fig:ablation_MIM}(b). The avgDist refers to the average difference between the predicted raw pixels and the nearest visible ones. A smaller avgDist indicates more accurate prediction. From the experimental results, we can observe that the avgDist for all masking methods increases smoothly with the rising of masking ratio. This is because as the masking ratio increases, the model needs to predict more information, leading to a higher probability of errors. However, at all masking ratios, the avgDist of our proposed TG-MIM is significantly lower than those of MAE and SimMIM. This indicates that the visual features extracted by the image encoder trained with TG-MIM contain richer visual details, thereby assisting in accurately reconstructing masked image regions.

\subsubsection{Analysis of Identity Supervised Global Visual Feature Calibration}
\label{subsubsec:ablation_ISGVFC}

\begin{figure}
\centering
\includegraphics[width=0.9\linewidth]{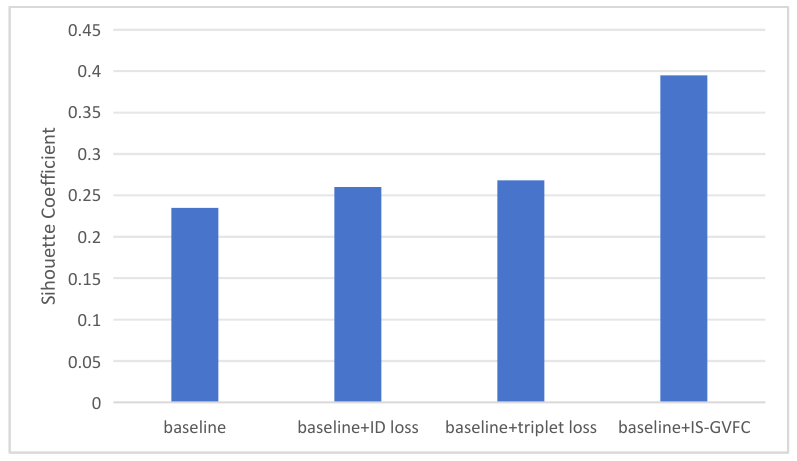}
\caption{Ablation study of global visual feature calibration.}
\label{fig:ablation_sih}
\end{figure}

\begin{figure}
\centering
	\subfloat[]{\includegraphics[width = 0.25\linewidth]{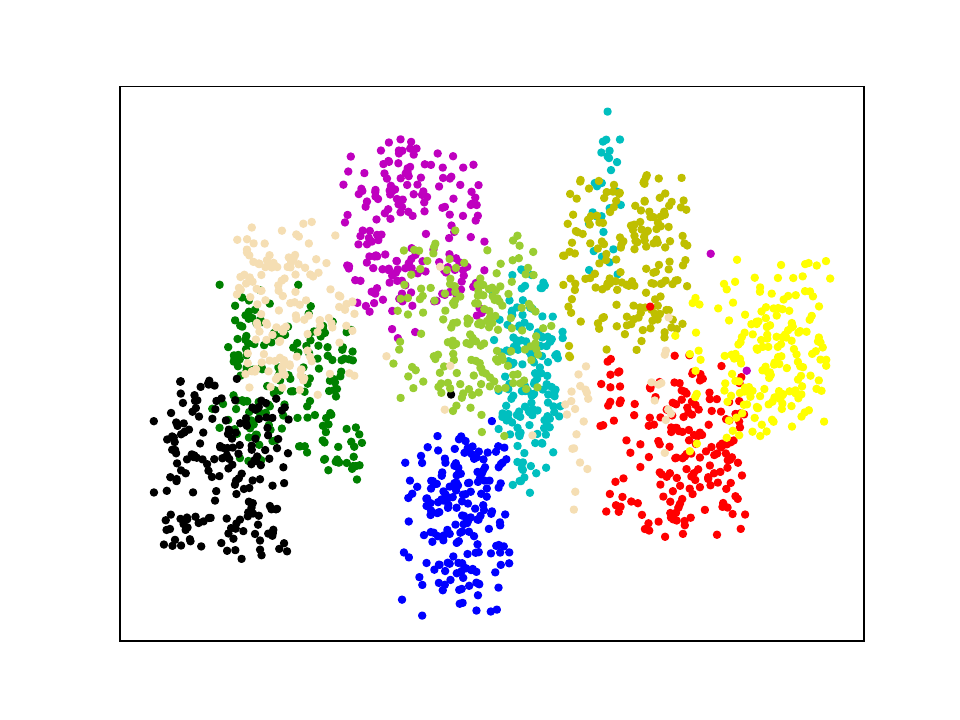}}
	\hspace{0.1mm}
	\subfloat[]{\includegraphics[width = 0.25\linewidth]{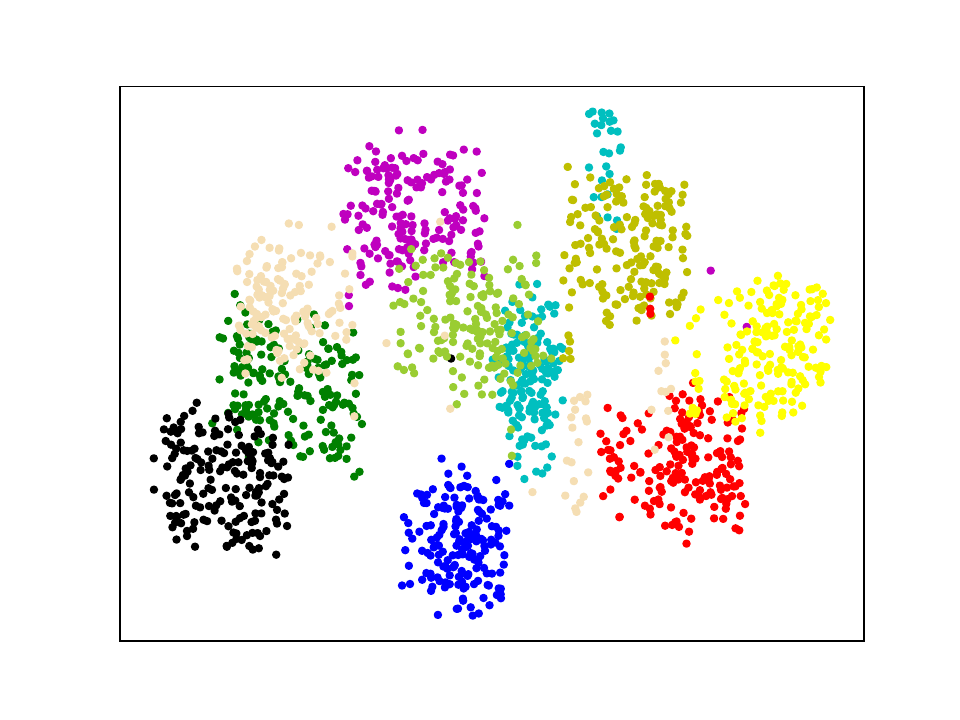}}
    \hspace{0.1mm}
	\subfloat[]{\includegraphics[width = 0.25\linewidth]{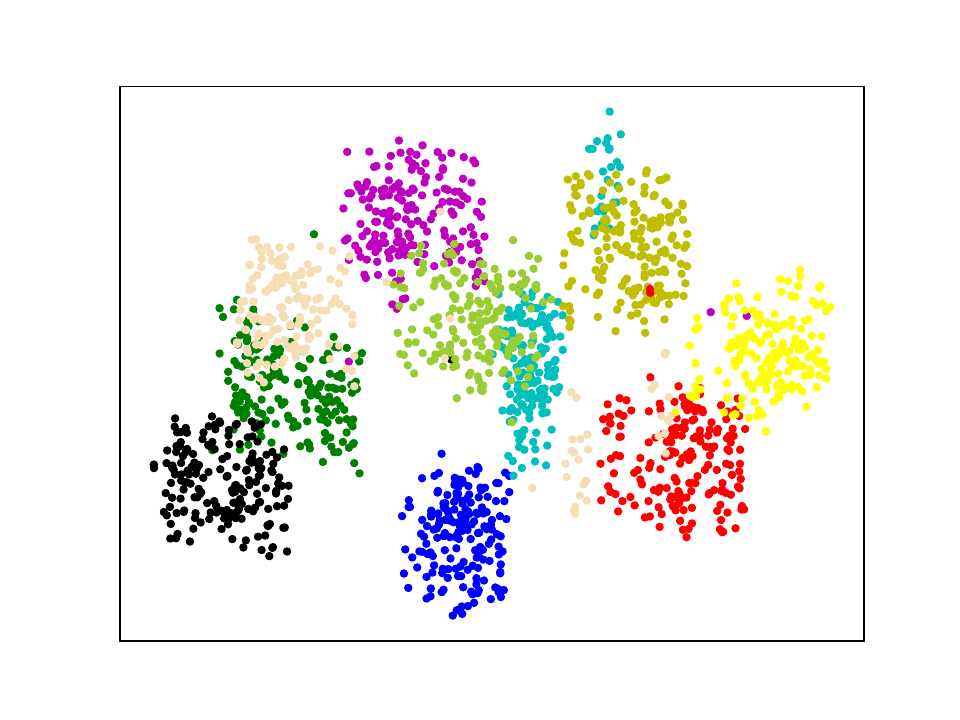}}
    \hspace{0.1mm}
	\subfloat[]{\includegraphics[width = 0.25\linewidth]{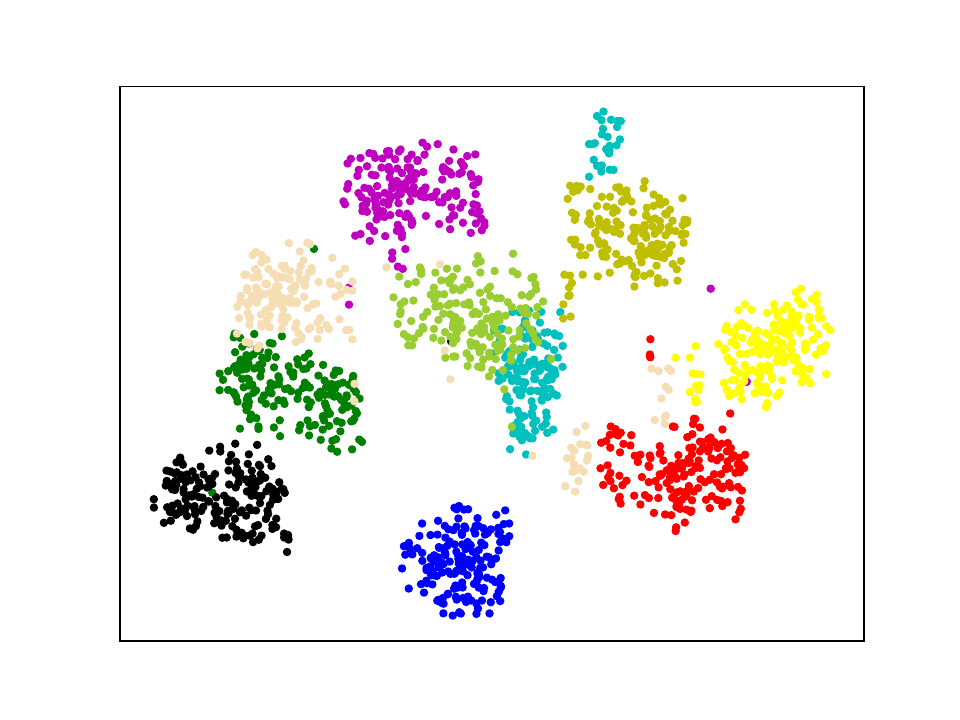}}
\caption{Visualization of global visual features under different calibration methods. Each color represents one identity class. (a) Global visual feature extracted by our model without any calibration. (b) Global visual feature extracted by our model guided by ID loss. (3) Global visual feature extracted by our model guided by triplet loss. (4) Global visual feature extracted by our model guided by our IS-GVFC.}
\label{fig:ablation_ISGVFC}
\end{figure}

In text-based person search task, there is a common issue of identity confusion as shown in Figure \ref{fig:motivation}(a). To alleviate this problem, we propose an Identity Supervised Global Visual Feature Calibration (IS-GVFC) auxiliary task as detailed in section \ref{subsubsec:ISGVFC} to guide the model to learn robust identity-aware global visual features. The ablation experiment results in Table \ref{tab:ablation_impact_VFE} have preliminarily demonstrated the effectiveness of the proposed method. In this section, we further quantitatively analyze the beneficial gains of IS-GVFC. Currently, existing methods commonly use ID loss or triplet loss to modulate image features, which involves utilizing the output of an image encoder for identity recognition or similarity ranking during training, aiming for the encoder to learn visual features that focus on pedestrian identities. Therefore, we compare the effect of our method with ID loss and triplet loss, as shown in Figure \ref{fig:ablation_sih}. Here, we use the silhouette coefficient \cite{R54} to quantify the effectiveness of visual feature calibration. The silhouette coefficient is a commonly used evaluation metric for assessing the quality of clustering, which combines both cohesion and separation factors. A higher silhouette coefficient value indicates that the calibrated features have a stronger resistance to identity confusion. The experimental results show that under the guidance of IS-GVFC, the global visual features have the highest silhouette coefficient. The commonly used ID loss and triplet loss constraints in existing methods can alleviate the issue of identity confusion to a certain extent, but their effectiveness is inferior to our IS-GVFC. To illustrate the impact of different calibration methods on the image encoder more vividly, we randomly sample pedestrians from 10 identity classes. We extract and project the global visual features using t-SNE \cite{R53} into a 2D space, as shown in Figure \ref{fig:ablation_ISGVFC}, which clearly demonstrates that, by using IS-GVFC, the visual features extracted by the image encoder exhibit better intra-ID cohesion and inter-ID dispersion. This, to some extent, explains why the retrieval accuracy of our proposed VFE-TPS model can surpass current SOTA models.

\begin{figure}
\centering
\includegraphics[width=\linewidth]{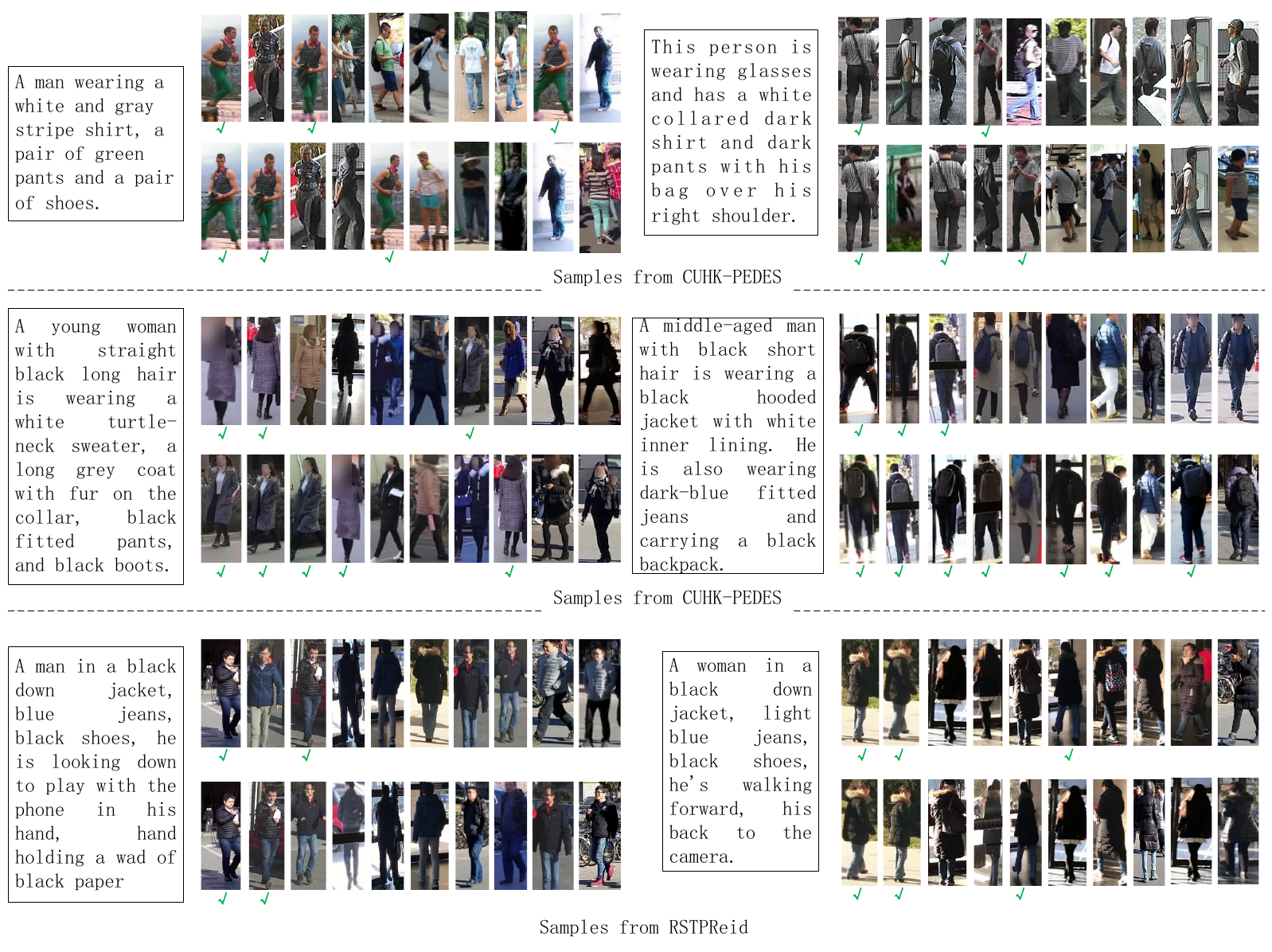}
\caption{Visualization of searching results. The first row of each sample shows the returned images by our model without TG-MIM and IS-GVFC, while the second row shows the returns of our full VFE-TPS model. The returned images are in order of similarity from high to low. The images with green check mark are the target pedestrians.}
\label{fig:visualization}
\end{figure}

\subsection{Visualization}
\label{subsec:visulization}
Figure \ref{fig:visualization} visualizes several examples from three datasets. The first row of each sample is returned by our model without TG-MIM and IS-GVFC, while the second row is returned by our full VFE-TPS model using the same query text. The target pedestrians are marked with green check at the bottom. As seen in Figure \ref{fig:visualization}, the baseline model is already able to retrieve the target pedestrians with a high probability, but it is prone to missing detections. When adding our proposed auxiliary tasks to the baseline model, this issue is alleviated and the model can retrieve the target pedestrians more accurately and comprehensively.

However, we also find that in the top-5 results returned by VFE-TPS, there are still non-target pedestrians present. For example, in the second example from RSTPReid, the pedestrians in the third and fourth images are not the target pedestrians. We notice that these two images are quite consistent with the query description, except for the additional single-shoulder bag worn by the women. This raises another problem of text-based person search task, that is, incomplete or vague query description. For such inadequately described query, the model can not accurately guess user's intent.

\section{Conclusion}
\label{sec:conclusion}
This paper proposes a Visual Feature Enhanced Text-based Person Search model (VFE-TPS) based on enhanced visual understanding of pedestrian images, which is considered crucial for improving the retrieval accuracy. We introduce the pre-trained CLIP model to extract basic features from images and texts. We also propose a text-guided masked image modeling auxiliary task that does not require additional annotation. By allowing the image encoder to predict masked image regions based on image context and query information, this task enhances the encoder's ability to understand visual details in the image. Furthermore, considering the issue of identity confusion in text-based person search task, we design an identity supervised global visual feature calibration auxiliary task to guide the model in learning identity-aware visual features. Experiments on multiple benchmarks validate the effectiveness of our proposed model.

However, we also find that our model returns non-target pedestrians when the query is vague or incomplete. To overcome this problem, it may be feasible to design a multi-round interactive model, which can return preliminary result according to the vague query and prompt the user to give more detailed description to further filter out the non-target pedestrians. This is also our future research content.

\section*{Acknowledgement}
This work was supported in part by the Natural Science Foundation of Zhejiang Province, China (Grant No. LQ21F020022) and China Postdoctoral Science Foundation (Grant No. 2022M720569).

\bibliography{refs}

\end{document}